\documentclass[conference]{IEEEtran}
\IEEEoverridecommandlockouts
\usepackage{cite}
\usepackage{amsmath,amssymb,amsfonts}
\usepackage{algorithmic}
\usepackage{graphicx}
\usepackage{textcomp}
\usepackage{xcolor}
\usepackage{booktabs}
 \usepackage{multirow}

\def\BibTeX{{\rm B\kern-.05em{\sc i\kern-.025em b}\kern-.08em
    T\kern-.1667em\lower.7ex\hbox{E}\kern-.125emX}}
\begin{document}

\title{SpheriGait: Enriching Spatial Representation via Spherical Projection for LiDAR-based Gait Recognition\\
\thanks{Identify applicable funding agency here. If none, delete this.}
}
\author {\IEEEauthorblockN {
Yanxi Wang \IEEEauthorrefmark {1}, ZhiGang Chang \IEEEauthorrefmark {2},Chen Wu \IEEEauthorrefmark {1},Zihao Cheng \IEEEauthorrefmark {1},Hongmin Gao \IEEEauthorrefmark {1}}
\IEEEauthorblockA {\IEEEauthorrefmark {1}Beijing Institute of Technology\\Email: hmgao@bit.edu.cn}
\IEEEauthorblockA {\IEEEauthorrefmark {2}Shanghai Jiao Tong University  }
}

\maketitle

\begin{abstract}
Gait recognition is a rapidly progressing technique for the remote identification of individuals. Prior research predominantly employing 2D sensors to gather gait data has achieved notable advancements; nonetheless, they have unavoidably neglected the influence of 3D dynamic characteristics on recognition. Gait recognition utilizing LiDAR 3D point clouds not only directly captures 3D spatial features but also diminishes the impact of lighting conditions while ensuring privacy protection. The essence of the problem lies in how to effectively extract discriminative 3D dynamic representation from point clouds. In this paper, we proposes a method named SpheriGait for extracting and enhancing dynamic features from point clouds for Lidar-based gait recognition. Specifically, it substitutes the conventional point cloud plane projection method with spherical projection to augment the perception of dynamic feature. Additionally, a network block named DAM-L is proposed to extract gait cues from the projected point cloud data. We conducted extensive experiments and the results demonstrated the SpheriGait achieved state-of-the-art performance on the SUSTech1K dataset, and verified that the spherical projection method can serve as a universal data preprocessing technique to enhance the performance of other LiDAR-based gait recognition methods, exhibiting exceptional flexibility and practicality.
\end{abstract}


\section{Introduction}
\label{sec:intro}
As a long-range, non-intrusive biometric technology, gait recognition boasts unique advantages in scenarios involving variations in lighting and changes in the subject’s attire\cite{chang4}. Its recognition holds immense potential for diverse real-world applications, encompassing security, criminal investigations, sophisticated home automation, human-computer interplay, and robotics, among others. Contrary to face recognition and person re-identification tasks that directly extract cues from RGB images, gait recognition often leverages posture keypoints, silhouette foregrounds, and other information to capture the motion characteristics of a target individual\cite{3D1,chang2,chang3}. This approach not only mitigates the effects of illumination changes and attire alterations but also diminishes the risks of personal privacy intrusion and leakage\cite{shallow,deep}.

Recent advancements in gait recognition using 2D sensor data have been remarkable in both controlled \cite{dataset1,dataset2,dataset3,chang1} and unstructured environments \cite{dataset4,dataset5,dataset6,3D1,lidar1}. Although cameras are commonly employed to capture 2D data, they fail to capture many 3D features, including viewpoints, shapes, and motion ranges\cite{3D1,dataset7}. Researchers have endeavored to reconstruct 3D representation (e.g., 3D meshes, skeletons, and depth maps) from 2D data or utilize multimodal techniques, yet these methods encounter challenges such as low resolution and poor illumination. Accurate 3D feature capture necessitates appropriate 3D data sensors. LiDAR technology can directly capture point clouds that encapsulate the 3D characteristics of all objects within a given space, rendering it widely applicable in fields such as autonomous vehicles, surveying and mapping, architecture, and engineering. Moreover, LiDAR offers remote sensing capabilities and remains impervious to lighting conditions and complex backgrounds\cite{Lidar3D4}, ensuring that 3D point clouds do not compromise the privacy of targets. The most pressing issue currently is determining the optimal method for processing 3D point clouds that preserves 3D dynamic features as much as possible while accounting for the correlation between points.

\begin{figure}[tbp]
\vspace{-0.3cm}
  \centering
  \includegraphics[width=8.5cm]{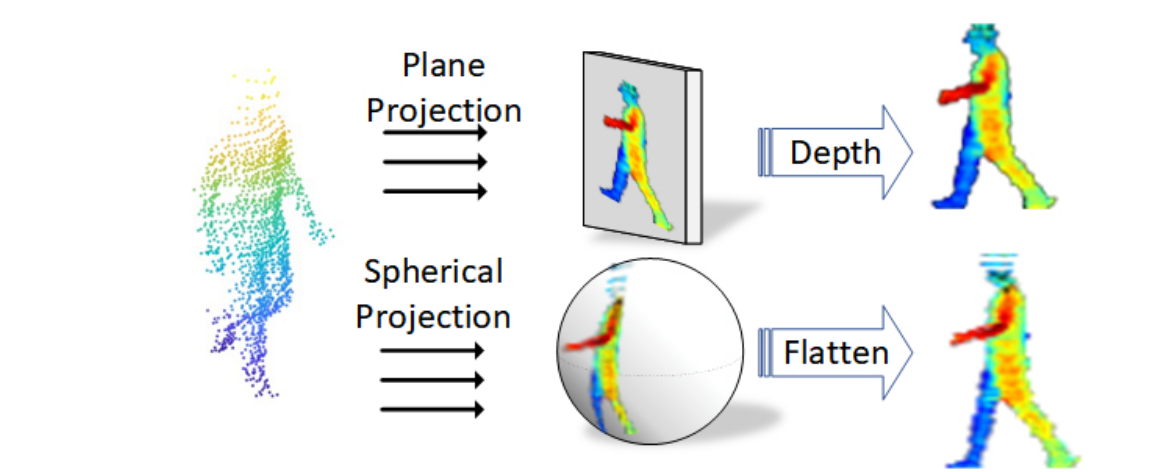}
  \caption{Plane projection and spherical projection.}
  \label{fig:fig1}
\vspace{-0.7cm}
\end{figure}

\begin{figure*}[th]
\vspace{-1.3cm}
  \centering
  \includegraphics[width=0.83\textwidth]{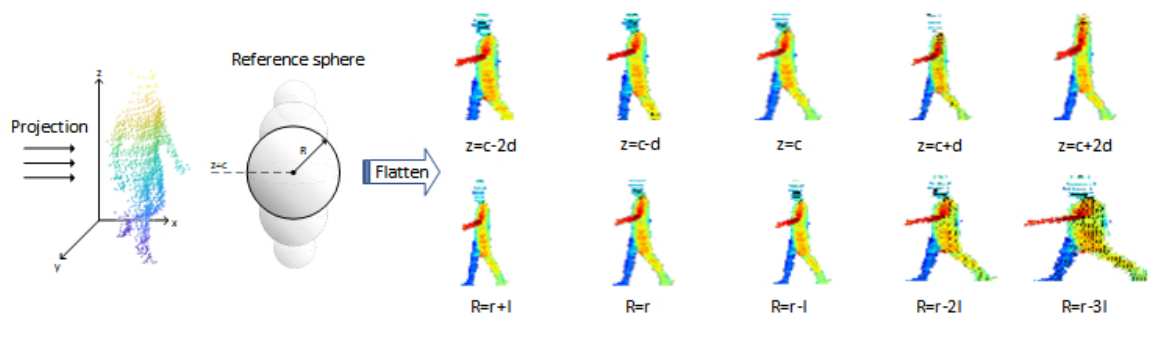}
  \caption{The influence of the z-axis coordinate and radius of the reference sphere on the projected depth map}
  \label{fig:fig2}
\vspace{-0.3cm}
\end{figure*}

Due to the unique format of 3D point clouds compared to image pixels and point-based gait recognition, two primary methods are currently used for feature recognition of 3D point clouds. 
The most common methods derive global context from sparse point clouds with limited local connectivity and directly perform identification based on the 3D point input\cite{Lidar3D2,Lidar3D3}.
However, the accuracy achieved by these methods is often lower than that obtained from camera-based silhouettes.
This may be due to an excessive focus on 3D features in these methods, while overlooking the interconnections between points\cite{lidar1}.
To extract fine-grained local information for greater accuracy, the second method, known as the projection method, is proposed. 
By projecting the 3D point cloud from the LiDAR range view into a depth image, it effectively captures fine-grained and distinctive gait features from the sparse point cloud.
\cite{Lidar3D1}suggests that the planar projection method may lose some dynamic features, prompting the use of a multimodal approach that combines both methods to further improve accuracy. 
However, this approach inevitably increases network complexity and the cost of practical applications. 
As shown in Fig.\ref{fig:fig1}, switching from planar to spherical projection of the point cloud can enhance the dynamic features of critical parts, such as the arms and legs in gait recognition, offering new perspectives for processing 3D point clouds.

To tackle this issue, we introduce a straightforward yet effective point cloud projection and gait recognition method, referred to as SpheriGait.
Specifically, SpheriGait first transforms the 3D point cloud in the LiDAR range view into a depth image using spherical projection, and then uses a convolutional network to extract gait features with 3D information highlighting the limb movements from the spherical projection.
By adjusting the radius and center position of the reference sphere, the proportion of low-impact features such as the torso and head is compressed, and the proportion of high-impact features such as the arms and back is increased in the depth map.
Extensive experiments demonstrate that (1) SpheriGait demonstrates that convolutional neural networks have the potential to be further improved in LiDAR gait recognition, (2) spherical projection is worthy of attention for its ability to enhance local features of 3D point clouds, and is expected to achieve success in other 3D recognition fields.

To summarize, our main contributions are as follows:
\begin{itemize}
\item We propose a gait recognition method named SpheriGait which employs spherical projection of LiDAR 3D point clouds to mitigate the issue of 3D feature loss due to point overlap in conventional point cloud planar projections.
\item SpheriGait propose a novel convolutional network block, DAM-L, for gait recognition, which can enhance the dynamic representation of spherical projections. 
\item SpheriGait achieved state-of-the-art performance on the previous LiDAR-based gait dataset (SUSTech1K), while also demonstrating the effectiveness of spherical projection in the original state-of-the-art methods, LidarGait and SwinGait. 
\end{itemize}
\vspace{-0.3cm}

\begin{figure*}[th]
\vspace{-1.3cm}
\begin{minipage}[t]{.68\linewidth}
  \centering
  \centerline{\includegraphics[width=14.0cm]{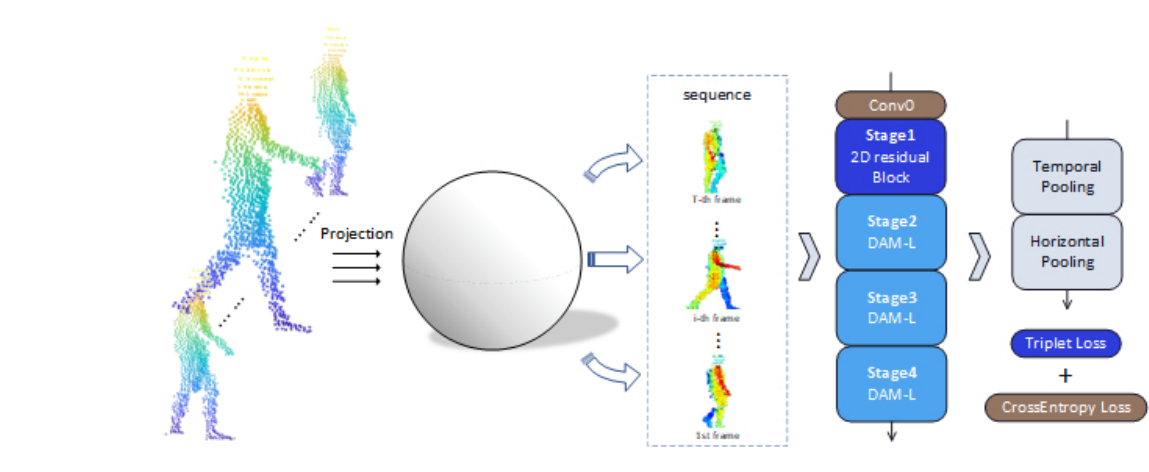}}
%  \vspace{1.5cm}
  \centerline{(a) Pipline}\medskip
\end{minipage}
\hfill
\begin{minipage}[t]{0.28\linewidth}
  \centering
  \centerline{\includegraphics[width=4.0cm]{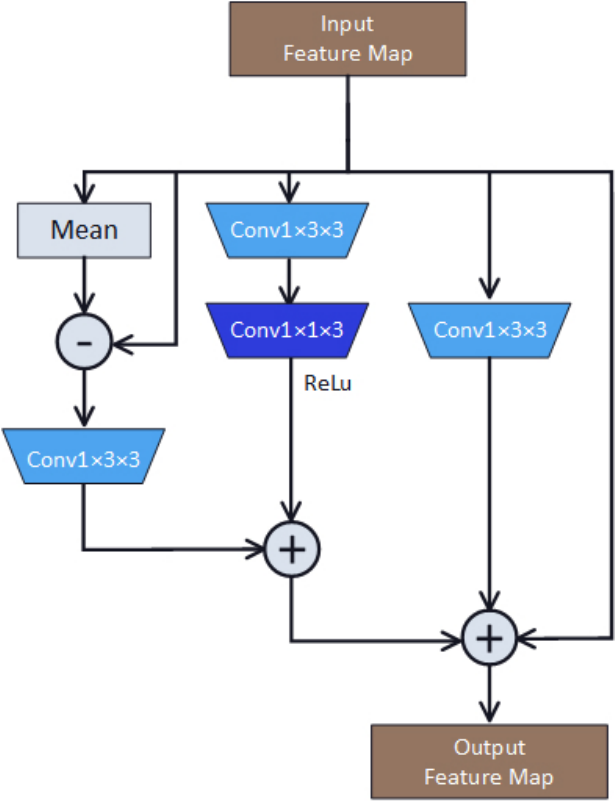}}
%  \vspace{1.5cm}
  \centerline{(b) DAM-L}\medskip
\end{minipage}
\caption{(a)We obtain gait depth maps that enhance dynamic features by projecting 3D point cloud spheres. Then, the feature map is obtained using four stages including three layers of DAM-L blocks. Finally, use Temporary Pooling and Horizontal Pooling Matching to extract features and calculate Triplet Loss and Cross Entropy Loss.(b)The DAM-L block utilizes two convolutional network branches to extract static and dynamic features, respectively.}
\label{fig:fig3}
\vspace{-0.3cm}
\end{figure*}

% \usepackage{multirow}
\begin{table*}[]
\centering
  {\small{
  \caption{Evaluation with different attributes on SUSTech1K valid + test set. We compare our method with silhouette-based SOTA method GaitBase, 3D point cloud-based SOTA method PointMLP and  PointGait,SOTA Multimodal method FreeGait and SOTA Projection-based LiDARGait.}
  \label{tab:constraction}
% \usepackage{multirow}
\begin{tabular}{c|c|c|cccccccc}
\hline
\multirow{2}{*}{Input}                                                     & \multirow{2}{*}{Methods} & \multirow{2}{*}{\begin{tabular}[c]{@{}c@{}}Overall\\ (Rank-1 acc)\end{tabular}} & \multicolumn{8}{c}{Probe Sequence (Rank-1 acc)}                                                                                                                                                                                                          \\ \cline{4-11} 
                                                                           &                          &                                                                                 & \multicolumn{1}{c|}{Normal} & \multicolumn{1}{c|}{Bag}   & \multicolumn{1}{c|}{Clothing}       & \multicolumn{1}{c|}{Carrying}       & \multicolumn{1}{c|}{Umberalla}     & \multicolumn{1}{c|}{Uniform}        & \multicolumn{1}{c|}{Occlusion} & Night \\ \hline
Silhouette                                                                 & GaitBase\cite{opengait}& 77.50                                                                           & \multicolumn{1}{c|}{83.09}  & \multicolumn{1}{c|}{79.34} & \multicolumn{1}{c|}{50.95}          & \multicolumn{1}{c|}{76.98}          & \multicolumn{1}{c|}{77.34}         & \multicolumn{1}{c|}{77.31}          & \multicolumn{1}{c|}{83.46}     & 26.65 \\ \hline
\multirow{2}{*}{\begin{tabular}[c]{@{}c@{}}3D Point\\  Cloud\end{tabular}} & PointMLP\cite{pointMLP}& 68.86                                                                           & \multicolumn{1}{c|}{76.03}  & \multicolumn{1}{c|}{71.91} & \multicolumn{1}{c|}{57.09}          & \multicolumn{1}{c|}{68.08}          & \multicolumn{1}{c|}{58.29}         & \multicolumn{1}{c|}{63.28}          & \multicolumn{1}{c|}{79.25}     & 70.75 \\ \cline{2-11} 
                                                                           & PointGait\cite{Lidar3D2}& 57.60                                                                           & \multicolumn{1}{c|}{68.63}  & \multicolumn{1}{c|}{65.15} & \multicolumn{1}{c|}{48.08}          & \multicolumn{1}{c|}{56.77}          & \multicolumn{1}{c|}{35.60}         & \multicolumn{1}{c|}{55.02}          & \multicolumn{1}{c|}{68.77}     & 61.70 \\ \hline
Multimodal                                                                 & HMRNet\cite{Lidar3D1}& 90.23                                                                           & \multicolumn{1}{c|}{92.71}  & \multicolumn{1}{c|}{92.34} & \multicolumn{1}{c|}{79.55}          & \multicolumn{1}{c|}{90.27}          & \multicolumn{1}{c|}{83.14}         & \multicolumn{1}{c|}{86.19}          & \multicolumn{1}{c|}{95.15}     & 90.35 \\ \hline
\multirow{3}{*}{Projection}                                                & CrossGait\cite{lidar3D6}& 84.90& \multicolumn{1}{c|}{90.60}   & \multicolumn{1}{c|}{-}     & \multicolumn{1}{c|}{71.20}           & \multicolumn{1}{c|}{-}              & \multicolumn{1}{c|}{66.50}          & \multicolumn{1}{c|}{-}              & \multicolumn{1}{c|}{-}         & 87.00\\ \cline{2-11} 
                                                                           & LiDARGait\cite{lidar1}& 86.77                                                                           & \multicolumn{1}{c|}{91.80}  & \multicolumn{1}{c|}{88.64} & \multicolumn{1}{c|}{74.56}          & \multicolumn{1}{c|}{89.03}          & \multicolumn{1}{c|}{67.50}         & \multicolumn{1}{c|}{80.86}          & \multicolumn{1}{c|}{94.53}     & 90.41 \\ \cline{2-11} 
                                                                           & \textbf{SpheriGait}        & \textbf{91.01}                                                                  & \multicolumn{1}{c|}{89.18}  & \multicolumn{1}{c|}{91.94} & \multicolumn{1}{c|}{\textbf{82.95}} & \multicolumn{1}{c|}{\textbf{90.73}} & \multicolumn{1}{c|}{\textbf{89.60}} & \multicolumn{1}{c|}{\textbf{91.53}} & \multicolumn{1}{c|}{94.17}     & 88.74 \\ \hline
\end{tabular}

}}

\end{table*}

\section{Method}

\subsection{spherical projection}
In this section, we present the LiDAR-based 3D point cloud format and its corresponding spherical projection method.
Fig.\ref{fig:fig1} illustrates the principles and distinctions between planar and spherical projections.
The 3D point cloud employed for gait recognition is obtained from the VLS128 LiDAR scanner, consisting of a collection of points that encapsulate 3D coordinate information.
The point cloud set can be expressed as $P=\{P^j_i|i=1,2,...,N;j=1,2,...,n_i\}$,where $N$ is the number of identities and $n_i$ is the sequence of each $i$-th identity.
Each point cloud sequence$P^j_i\in \mathbb{R}^{T*N*C}$,is with $T$ frames and $N$ points for each frame,where $C$ represents the number of feature channels.
For the coordinates of a 3D point  $p=(x.y,z)^T$ in the dataset $P$.
In plane projection, since the laser radar uses a cylindrical coordinate system to collect point sets, the horizontal and vertical coordinates of the corresponding depth map are$arctan(x,y)$ and $arcsin(z,\sqrt{x^2+y^2+z^2})$.
The so-called spherical projection is to project the point cloud horizontally onto a reference sphere outside the point cloud area, and then flatten the sphere back to the plane coordinate system, as shown in Fig.\ref{fig:fig1} .
Therefore, we need to modify the function for plane projection and use the following spherical projection function:
\begin{equation}
\begin{split}
    h=arccos(y,R)/\Delta\theta  \\
    v=arctan((z-z_r),R)/\Delta\phi
\label{eq:3-1} 
\end{split}
\end{equation}
The point $p$ is mapped to its corresponding 2D pixel coordinates $(h,v)$ in the depth map based on spherical projection.
The $\Delta\theta$ and$\Delta\phi$ represent the average resolution of the horizontal and vertical angles between successive beam emitters.
The radius $R$ and center height $z_r$ of the reference sphere determine the density of each region within the depth map.
Fig.\ref{fig:fig2} shows the projection result of different $r$ and $z_r$.where $r=\sqrt{x^2+y^2+(z-c)^2}$ is the average distance between the point set and the origin within the range of  LiDAR, $l$ and $d$ are the change steps, and $c=(z_{max}+z_{min})/2$ is the center height of the $z$ coordinate of the projection area.
Each element in the map at position$ (h, v)$ is filled with $D$, where $D=\sqrt{x^2+y^2}$.
The depth projection is subsequently normalized and converted from single-channel images into RGB images.

\vspace{-0.3cm}

\subsection{Pipline}
Spherical projection effectively accentuates the proportion of dynamic gait features within the data, prompting us to employ a convolutional neural network that excels in extracting dynamic features to enhance the accuracy of gait recognition.
\cite{dygait} proposed a block DAM for generating dynamic feature maps by utilizing the differences between gait features of each frame and gait templates based on mean functions, and on this basis, SpheriGait replaced the convolutional network that performed poorly on depth maps to establish the Dynamic Augmentation Module in LiDAR (DAM-L) suitable for point cloud spherical projection.
In SpheriGait, we employ a convolutional neural network to process the depth map derived from the spherical projection of the point cloud, as illustrated in the pipeline shown in the Fig.\ref{fig:fig3}(a).
The Backbone Network of this method consists of four stages, with the first stage being a 2D residual unit and the remaining three stages being DAM-L blocks.

As shown in Fig.\ref{fig:fig3}(b), DAM-L blocks has two branch tasks in total.
The first branch aims to capture the spatiotemporal representation of gait dynamic features by passing the input feature map through two networks with differently sized convolution kernels, then adding the result to the output of the input feature map, which has undergone frame dimension mean subtraction and passed through a convolutional network.
The second branch is designed to capture spatiotemporal representations of the static aspects of gait, which are extracted from the input feature map through a convolutional network layer. 
Finally, the two branches are combined with the input features to produce the output feature map of the DAM-L module.

Upon completing the four stages, the feature map is directed into the Temporal Pooling and Horizontal Pooling layers to extract features essential for recognition.
To enhance performance, our network undergoes training utilizing both triplet loss and cross-entropy loss.

\section{Experiments}
\label{sec:typestyle}
\vspace{-0.1cm}
\subsection{Comparison with SOTA Methods}
Table.\ref{tab:constraction} presents a comparison between our method and state-of-the-art (SOTA) gait recognition techniques on the SUSTech1K datasets.
Our approach surpasses the existing SOTA method, LidarGait \cite{lidar1}, in single depth map recognition, achieving a 4.42\% improvement in Rank-1 accuracy on the SUSTech1K dataset.
These achievements stem from the feature enhancement from spherical projection and the dynamic extraction capability of convolutional networks in SpheriGait.
Compared with the existing SOTA method that combines depth maps with 3D point clouds and other multimodal recognition method\cite{Lidar3D1,Lidar3D5}, we also achieved a 0.78\% improvement in Rank-1 accuracy.
This demonstrates that the projection method also has the potential to extract dynamic 3D features, and complex multimodal recognition is not necessary.
These point-based methods try hard to handle sparse representation\cite{Lidar3D2,pointMLP}, but due to the disorder of 3D point clouds, the results are generally inferior to silhouette-based methods that lose 3D features\cite{opengait,deep,lidar3D6}.
In contrast, our SpheriGait successfully captures 3D dynamic features and explicitly models gait-related movements, leading to more robust performance, even in challenging scenarios.

\subsection{Dataset}\label{AA}
\textbf{SUSTech1K }\cite{lidar1}serves as the primary benchmark for studying 3D feature-based gait recognition, being the first publicly available dataset to encompass a wide range of conditions.
It is also the only public dataset that presents gait expressions in the wild through LiDAR 3D point clouds, forming the basis for the research in this study. 

\subsection{Ablation for Spherical Projection}
Fig.\ref{fig:fig2} demonstrates that altering the position and size of the reference sphere greatly influences the depth map in spherical projection. 
Thus, to identify the optimal spherical projection method for gait recognition, we incrementally adjust the z-axis coordinate and the radius of the reference sphere. 

\begin{figure}[t]
\vspace{-0.2cm}
\begin{minipage}[t]{0.48\linewidth}
  \centering
  \centerline{\includegraphics[width=4.6cm]{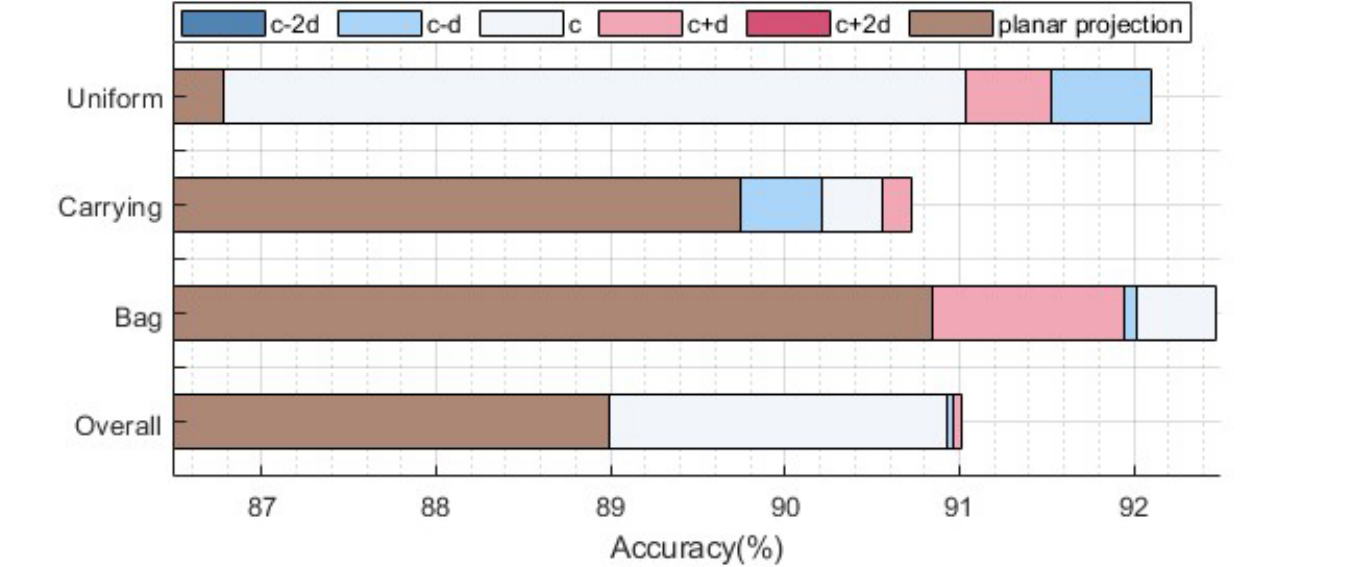}}
%  \vspace{2.0cm}
  \centerline{(a)}\medskip
\end{minipage}
\begin{minipage}[t]{0.48\linewidth}
  \centering
  \centerline{\includegraphics[width=4.6cm]{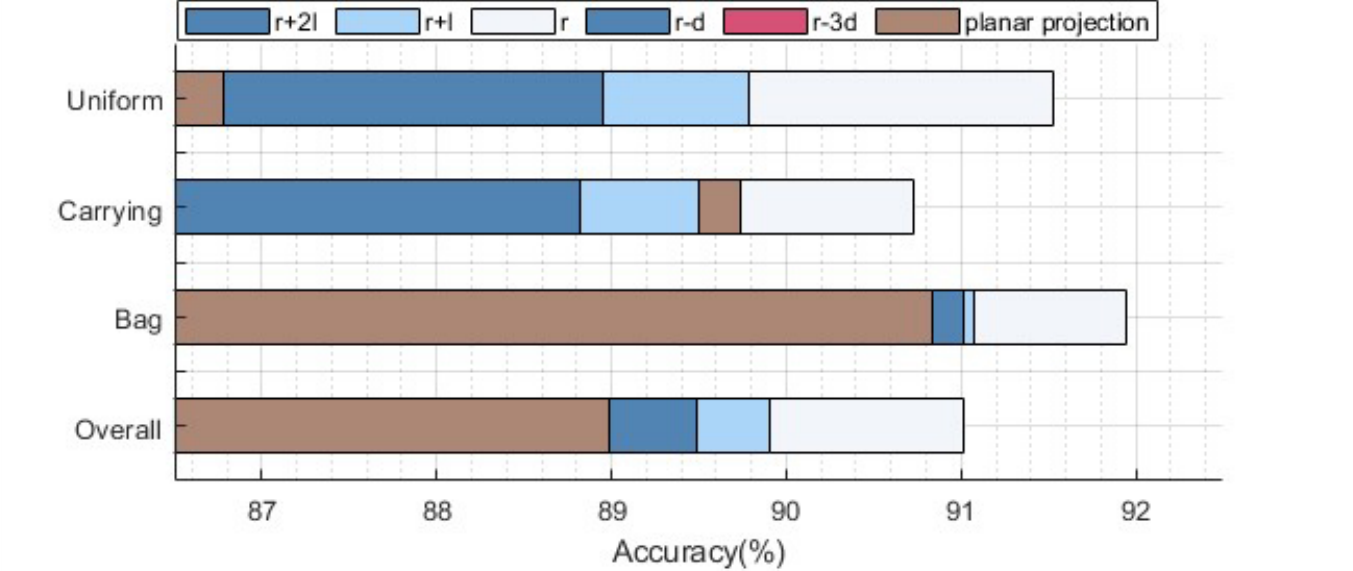}}
%  \vspace{1.5cm}
  \centerline{(b)}\medskip
\end{minipage}

\caption{The influence of reference spheres with varying (a) z-axis coordinates and (b) radii on the identification performance of SpheriGait. The figure displays the recognition accuracy of plane projection and the top three reference sphere projections in each category, with both achieving optimal accuracy at positions slightly above the center height and slightly below the average radius, respectively.}
\label{fig:fig4}
\end{figure}

As shown in Fig.\ref{fig:fig2} the z-axis height of the reference sphere defines the center of compression, while its radius determines the intensity of that compression.  
Fig.\ref{fig:fig4} illustrates the optimal z-axis height and radius for achieving the highest overall accuracy. 
Nevertheless, it is evident that this reference sphere configuration does not achieve the highest accuracy in all scenarios.
Consequently , in practical applications, the choice of the projection reference sphere must be tailored to real-world conditions.

\subsection{Ablation for DAM-L}
\cite{dygait} utilizes a 3*3*3 convolution kernel in dynamic feature extraction tasks in DAM block, with each block comprising two layers of the sum of branch tasks.
SpheriGait uses pseudo 3D residual units and 2D convolution kernels to replace the 3D convolution kernels in DAM to obtain the 3D convolution layer in the DAM-L block shown in Fig.\ref{fig:fig3}(b), and only uses one layer for branch summation.
Table.\ref{tab:DAM} compares the impact of different blocks on recognition accuracy, revealing that single-layer DAM-L performs significantly better than DAM.
This indicates that for depth maps with enhanced dynamic features, pseudo 3D residual units outperform the 3D convolution layer.

\begin{table}[t]
\centering
  {\small{
  \caption{Comparison of accuracy using single-layer versus double-layer DAM or DAM-L blocks for recognition shows that single-layer DAM blocks achieve the best performance. }
  \label{tab:DAM}
% \usepackage{multirow}

\begin{tabular}{c|cc}
\hline
\multirow{2}{*}{Type} & \multicolumn{2}{c}{\begin{tabular}[c]{@{}c@{}}Overall\\ (Rank-1 acc)\end{tabular}} \\ \cline{2-3} 
                      & \multicolumn{1}{c|}{1 layer}                          & 2 layers                   \\ \hline
DAM                   & \multicolumn{1}{c|}{76.66}                            & 50.75                      \\ \hline
DAM-L                 & \multicolumn{1}{c|}{\textbf{91.01}}                   & 74.66                      \\ \hline
\end{tabular}

}}

\end{table}

\subsection{Comparison of projection methods}

\begin{figure}[t]
\vspace{-0.2cm}
\begin{minipage}[t]{0.48\linewidth}
  \centering
  \centerline{\includegraphics[width=4.5cm]{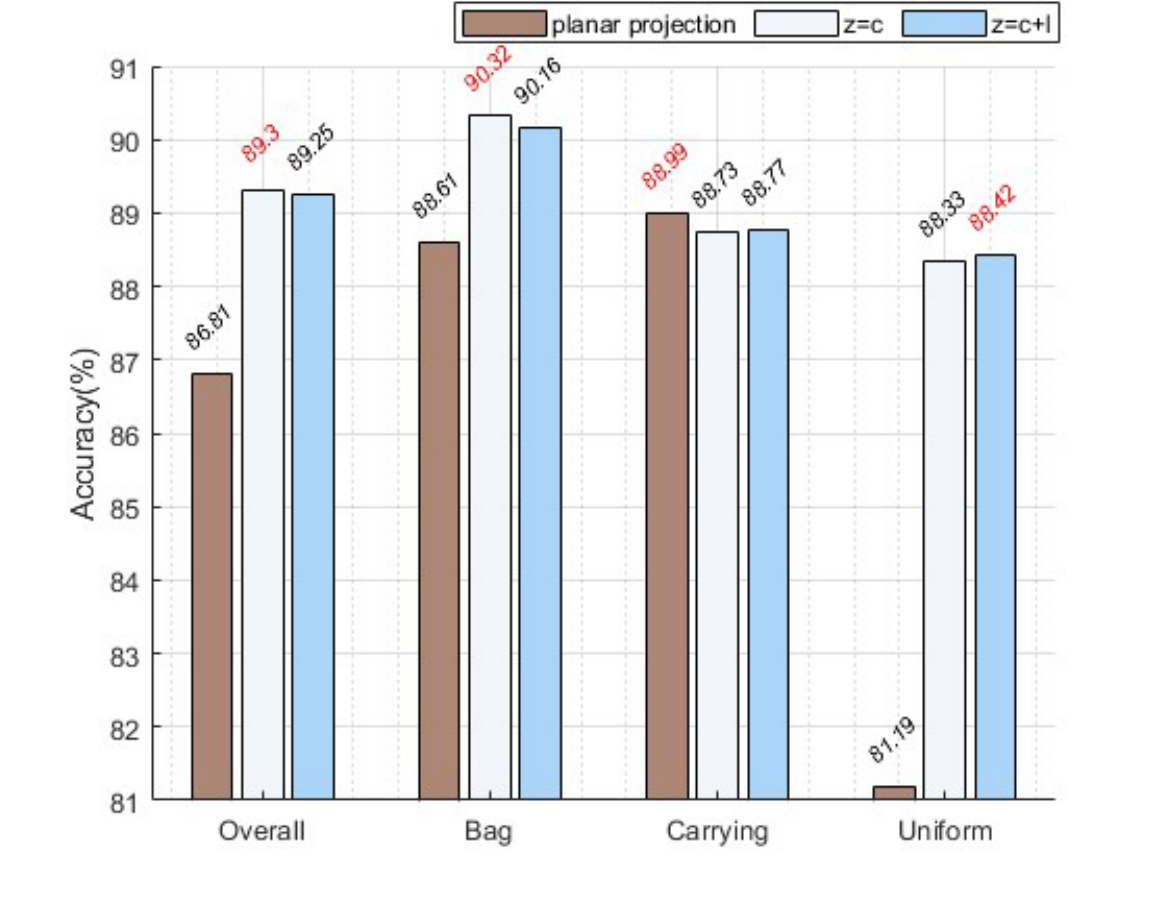}}
%  \vspace{2.0cm}
  \centerline{(a)}\medskip
\end{minipage}
\begin{minipage}[t]{0.48\linewidth}
  \centering
  \centerline{\includegraphics[width=4.5cm]{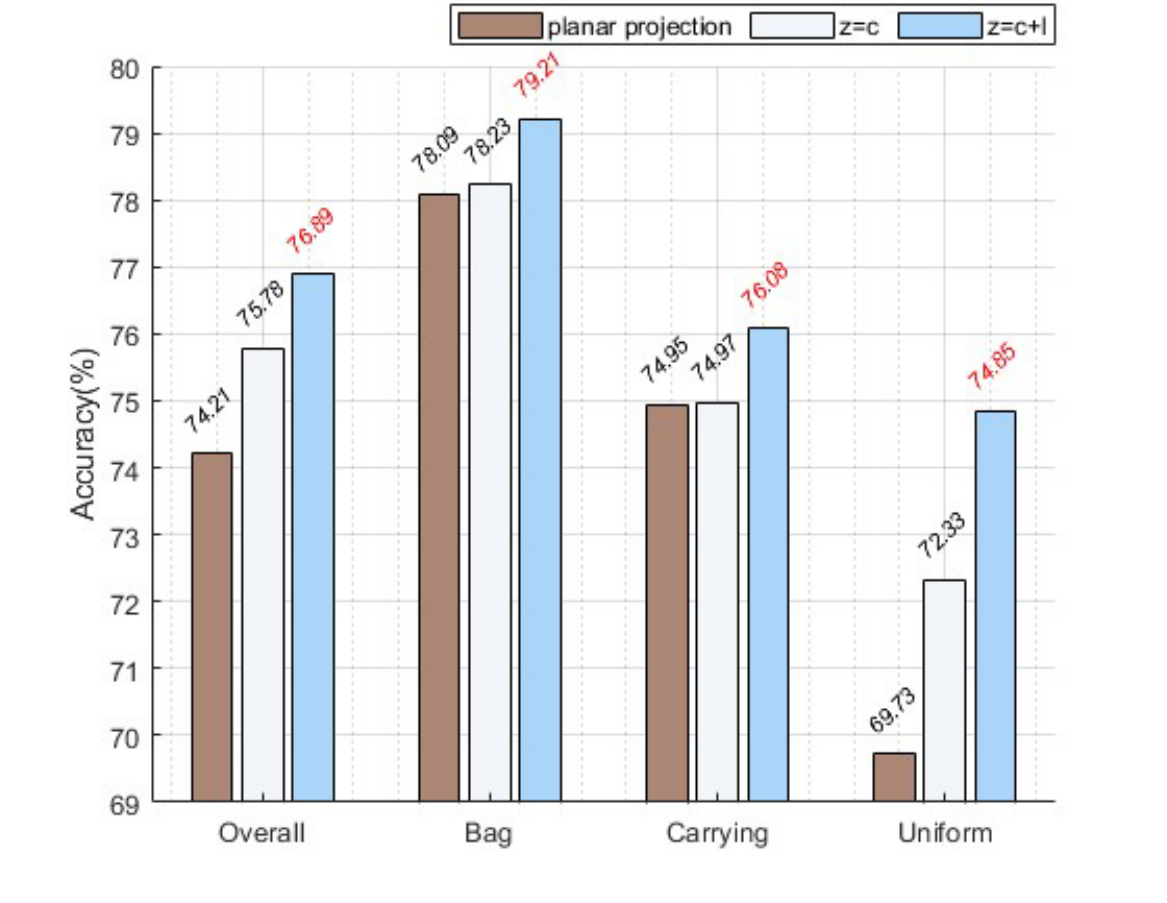}}
%  \vspace{1.5cm}
  \centerline{(b)}\medskip
\end{minipage}

\caption{(a) Comparing the accuracy of spherical projection and planar projection under the same convolutional network recognition method, it can be found that the spherical projection of the reference sphere $z=c$ has the best overall accuracy.(b) Under the same Teansformer recognition method, it can be found that the spherical projection of the reference sphere $z=c+l$ has the best overall accuracy.}
\label{fig:fig5}
\end{figure}

To demonstrate the advantages of spherical projection, we compare its performance against existing SOTA convolutional network and Transformer methods LidarGait and SwinGait for planar projection of LiDAR\cite{deep,lidar1}. 
Fig.\ref{fig:fig5}(a) shows that the best results were achieved when the reference sphere was located at $z=c$, the spherical projection outperforms LidarGait in \textbf{Overall (2.80\%), Bag (1.68\%), Clothing (1.91\%) and Uniform(7.44\%)} by a large margin,  especially in \textbf{Umbrella (21.31\%)}.

The visual Transformer method is a widely adopted recognition technique based on self-attention. 
Fig.\ref{fig:fig5}(b) compares the recognition accuracy of various projection methods under the Transformer method SwinGait\cite{deep}, showing that the maximum accuracy is achieved when the reference sphere is located at $z=c+l$.
This is because sphere projection enhances dynamic features, demonstrating that even in challenging scenarios and recognition methods, it can yield more robust performance.

\vspace{-0.1cm}

\section*{Conclusion}

In this paper, we propose a novel method for gait recognition of 3D point clouds using LiDAR. 
The proposed method SpheriGait incorporates a point cloud spherical projection technique and a Dynamic Augmentation Module in LiDAR (DAM-L) block for extracting dynamic features. 
Spherical projection effectively enhances dynamic features in point clouds, while DAM-L blocks are designed to extract these features. 
Comparative experiments on LidarGait and SwinGait demonstrate that the proposed spherical projection method can achieve optimal dynamic feature enhancement.

\vspace{12pt}

\end{document}